\documentclass[runningheads]{llncs}

\usepackage{eccv}

\usepackage{eccvabbrv}

\usepackage{graphicx}
\usepackage{booktabs}
\usepackage{float}
\usepackage{pifont}

\usepackage[accsupp]{axessibility}  

\usepackage{hyperref}

\usepackage{orcidlink}

\begin{document}

\title{JEPADepth: Masked Predictive Representation Learning for Self-Supervised Monocular Depth Estimation} 

\titlerunning{JEPADepth}

\author{Ionuț Grigore\inst{1}\orcidlink{0009-0004-5006-0682} \and
Călin-Adrian Popa\inst{1}\orcidlink{0000-0003-4445-8091}}

\authorrunning{I. Grigore and C.-A. Popa}

\institute{Politehnica University of Timișoara, Romania
\email{\{ionut.grigore,calin.popa\}@cs.upt.ro} \\
\vspace{10pt}
\href{https://github.com/ionut-grigore99/JEPADepth}{\textcolor{blue}{https://github.com/ionut-grigore99/JEPADepth}}
}

\maketitle

\begin{abstract}
Self-supervised monocular depth estimation typically relies on photometric
reconstruction losses that couple depth, pose, and appearance assumptions.
In this paper, we propose JEPADepth, a self-supervised monocular depth framework that incorporates a complementary training objective inspired by Image Joint-Embedding Predictive Architectures (I-JEPA) for self-supervised depth learning. Our method augments a standard photometric pipeline with a masked prediction loss computed in the representation space
of a pretrained DINOv3 Vision Transformer encoder. A predictor infers target-region embeddings from visible context-region embeddings under structured masking, and is discarded along with the target encoder at inference time, adding no deployment cost. On KITTI, adding the JEPA objective consistently improves performance over the same DINOv3-based photometric baseline, without changing the inference-time architecture.
Compared to prior monocular self-supervised methods, JEPADepth is competitive with state-of-the-art transformer-based approaches and outperforms strong CNN-based baselines on the standard benchmark. In zero-shot transfer (trained on KITTI and evaluated without fine-tuning), JEPADepth achieves the best or near-best performance among the compared methods on both Make3D and Cityscapes across multiple metrics.

\end{abstract}

\section{Introduction}
\label{sec:intro}

Monocular depth estimation is a core capability for 3D scene understanding, enabling applications in robotics, autonomous driving, and augmented reality \cite{tang2022perception}. While supervised learning approaches achieve strong accuracy, they require dense ground-truth depth that is expensive to obtain and often limited to specific sensors and environments. This has motivated a large body of work on self-supervised monocular depth estimation, where depth is learned directly from unlabeled videos by enforcing geometric consistency across adjacent frames.

The dominant self-supervised paradigm relies on view synthesis: a depth network and a pose network are trained such that a target frame can be reconstructed by warping source frames using the predicted depth and camera motion \cite{zhou2017unsupervised}. The photometric reconstruction error provides the main learning signal, typically complemented by smoothness and masking heuristics to mitigate occlusions and moving objects \cite{godard2019digging}. Despite its success, photometric supervision is inherently sensitive. It depends on assumptions such as brightness constancy and sufficient texture, and it can be disrupted by reflective surfaces \cite{shi20233d, choi2025self}, illumination changes \cite{vankadari2023sun}, and non-rigid motion \cite{zhao2020monocular}. Moreover, because the loss is computed in pixel space, optimization can encourage solutions that memorize appearance statistics rather than enforcing higher-level scene structure \cite{li2021sins}, which can limit robustness and cross-dataset transfer \cite{saunders2023self}.

In parallel, self-supervised representation learning has recently produced powerful visual encoders that capture semantic and structural regularities without manual labels. Vision Transformers \cite{dosovitskiy2020image} pretrained with large-scale objectives, such as DINOv3 \cite{simeoni2025dinov3}, learn patch-level features that align with object boundaries, layout, and part structure. While these static features are highly effective for downstream tasks, depth estimation fundamentally requires reasoning about spatial geometry and occlusion. This suggests an opportunity: instead of relying solely on pixel reconstruction, self-supervised depth estimation can be regularized through feature-space predictive objectives that force the model to understand spatial relationships.

Motivated by this, we introduce JEPADepth, a self-supervised depth framework that integrates an Image Joint-Embedding Predictive Architecture (I-JEPA) \cite{assran2023self} objective as a complementary training signal. Our approach augments a standard photometric depth pipeline with a masked predictive loss computed on patch tokens produced by a DINOv3 encoder. Given structured context and target masks on the patch grid, a predictor is trained to infer target-region embeddings from the visible context-region embeddings. Rather than reconstructing pixels, the JEPA-style objective encourages region-level consistency and predictive understanding in representation space. The depth network is trained end-to-end using a combination of the classical photometric loss and the JEPA representation prediction loss, yielding a model that is simultaneously constrained by multi-view geometry and regularized by predictive feature learning.

A central motivation of this design is to improve robustness beyond the training
distribution. Photometric self-supervision can be sensitive to appearance changes
(e.g., illumination, texture, and camera characteristics), which can degrade
cross-dataset transfer. We therefore add a masked prediction objective in
representation space as an auxiliary training signal. Empirically, on KITTI we
observe improved in-domain accuracy when adding the JEPA loss to the same DINOv3-based
photometric model. Moreover, a model trained on KITTI
shows stronger zero-shot results on Make3D and Cityscapes without fine-tuning, indicating improved transfer under
this evaluation protocol.

Our main contributions are as follows:
\begin{itemize}
\item \textbf{JEPA-regularized self-supervised depth estimation}. We propose integrating an I-JEPA-inspired masked representation prediction objective into a standard self-supervised monocular depth pipeline based on photometric view synthesis, leveraging a pretrained DINOv3 encoder to perform predictions directly in representation space rather than pixel space.
\item \textbf{Improved performance over a strong baseline}. We show that adding the I-JEPA loss to a DINOv3-based photometric depth framework yields better results than training with photometric supervision alone. Because the predictor and target encoder are discarded after training, these gains are achieved without adding computational overhead during inference.
\item \textbf{Enhanced generalization under domain shift}. With training on KITTI only,
we report zero-shot evaluation on Cityscapes and Make3D without fine-tuning, where the
proposed objective improves results under these benchmarks.
\end{itemize}

\section{Related Work}

\subsection{Self-Supervised Depth Estimation}
In the absence of ground-truth depth annotations, self-supervised monocular 
depth estimation methods exploit geometric consistency in visual data. 
SfMLearner~\cite{zhou2017unsupervised} established the dominant paradigm: 
jointly training a depth and pose network using photometric reconstruction loss. 
Subsequent works introduced robust losses~\cite{shu2020feature,gordon2019depth}, 
feature-level consistency~\cite{zhan2018unsupervised}, auxiliary 
signals~\cite{watson2019self}, and strategies for dynamic 
objects~\cite{godard2019digging,ranjan2019competitive,casser2019unsupervised}.
Transformer-based encoders further improved accuracy~\cite{zhao2022monovit, zhang2023lite}, 
and more recently structured state space models have been explored for 
long-range dependency modeling~\cite{grigore2024mambadepth}.

Closest to our work are feature-space self-supervised methods. Zhan et 
al.~\cite{zhan2018unsupervised} and Shu et al.~\cite{shu2020feature} augment the 
photometric pipeline with a cross-frame feature reconstruction loss and a 
feature-metric warping loss, respectively. In both cases the auxiliary signal is a 
consistency between features extracted from two different frames, such as a stereo 
pair or a photometrically warped view. In contrast, our objective operates entirely 
within a single image through structured masking and a stop-gradient EMA target 
encoder, making it independent of the multi-view photometric objective rather than 
a reformulation of it. To our knowledge, JEPADepth is the first method to use an 
I-JEPA masked predictive objective as an online auxiliary training signal inside a 
photometric self-supervised depth pipeline.

A related line of work introduces masked prediction as a regularizer for 
self-supervised depth. MIMDepth~\cite{chawla2023image} adds a pixel-space 
reconstruction loss alongside photometric supervision. Unlike this approach, 
JEPADepth performs prediction in representation space following 
I-JEPA~\cite{assran2023self}, avoiding sensitivity to pixel-level appearance.

\begin{figure}[tb]
  \centering
  \includegraphics[height=7.5cm]{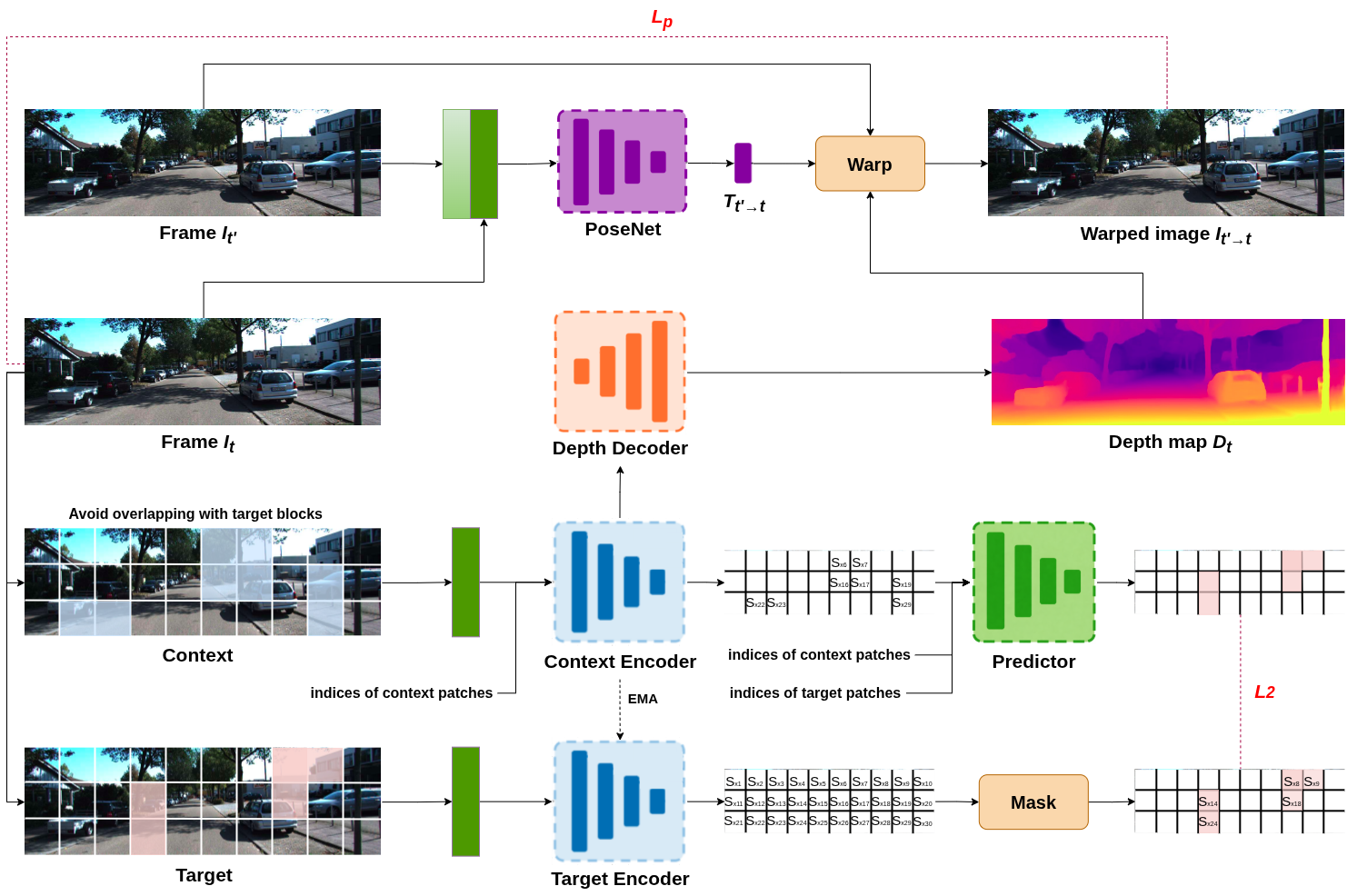}
  \caption{\textbf{Overview of JEPADepth}. A DINOv3 ViT encoder extracts patch tokens from the target image. A context mask selects visible tokens (blue) that are fed to a predictor, which generates embeddings for multiple masked target regions (red). In parallel, an EMA-updated target encoder produces target-region embeddings used as stop-gradient learning targets. The predicted and target embeddings are matched with a JEPA loss, which is combined with the standard photometric view-synthesis loss to train the depth network end-to-end.
  }
  \label{fig:JEPADepth}
\end{figure}

\subsection{Joint Embedding Predictive Architecture}
I-JEPA~\cite{assran2023self} is a self-supervised framework that predicts latent 
representations of masked image regions from visible context patches, rather than 
reconstructing pixels. A context encoder processes visible patches while a 
momentum-updated (EMA) target encoder provides stop-gradient prediction targets. 
This design captures structured spatial dependencies without pixel-level 
supervision. The JEPA framework has since been extended to 
video~\cite{assran2025v}, 3D point clouds~\cite{saito2025point}, and world 
models~\cite{terver2025drives}. JEPADepth does not propose a new JEPA 
architecture but uses the I-JEPA objective as a training signal on top of a 
standard photometric pipeline, a combination not explored in prior work.

\begin{figure}[t]
  \centering
  \includegraphics[height=6cm]{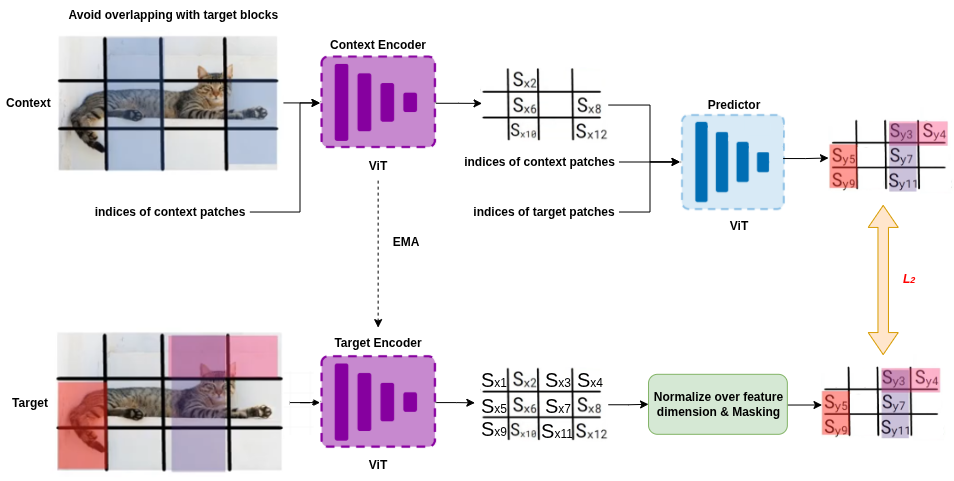}
  \caption{\textbf{Overview of the Image-JEPA training framework}~\cite{assran2023self}.
  Visible context patches are processed by a context encoder, while masked target
  patches are encoded by a momentum-updated target encoder (EMA). A predictor
  maps context representations to the latent representations of target patches.
  The model is trained by minimizing a regression loss in feature space between
  predicted and target embeddings after normalization and masking.}
  \label{fig:ijepa}
\end{figure}

\section{Method}
\subsection{Self-Supervised Depth Estimation Framework}
Our framework is built upon Monodepth2~\cite{godard2019digging} and follows the
standard Structure from Motion (SfM) self-supervised paradigm, where a monocular
camera moves through a rigid environment to provide multiple views of the same scene.
We include the full derivation here for completeness.
 
\subsubsection{Notation and View Synthesis}
 
Let $I_{t}\in\mathbb{R}^{H\times W\times3},\;t\in\{-1,0,1\}$ be a frame in a monocular
video sequence captured by a moving camera, where $t$ is the frame time index.
Let $D_{t}\in\mathbb{R}^{H\times W}$ denote the depth map corresponding to image $I_{t}$.
The camera pose change from time $0$ to time $t\in\{-1,1\}$ is encoded by a $3\times3$
rotation matrix $R_{t}$ and a translation vector $\mathbf{t}_{t}$.
We obtain the $4\times4$ camera transformation matrix:
\begin{equation}
M_{t}=\begin{bmatrix}R_{t} & \mathbf{t}_{t}\\ 0 & 1\end{bmatrix}.
\label{eq:transform}
\end{equation}
 
Our aim is to train two networks to simultaneously estimate the pose of the camera and
the structure of the scene:
\begin{equation}
M = \theta_{\text{pose}}(I_{t}), \qquad D = \theta_{\text{depth}}(I_{t}).
\label{eq:networks}
\end{equation}
 
Self-supervised depth prediction reformulates the learning task as a novel view-synthesis
problem. During training, the coupled network synthesizes the photometric appearance of a
target frame from a source frame viewpoint, using the depth map as an intermediate
variable.
 
\subsubsection{Perspective Projection and Backprojection}
 
Let $(u,v)\in\mathbb{R}^{2}$ be the calibrated pixel coordinates in image $I_{0}$,
with origin $(0,0)$ at the top-left. A 3D point $(X,Y,Z)\in\mathbb{R}^{3}$ projects
onto $(u,v)$ through the perspective projection operator:
\begin{equation}
\pi(X,Y,Z)=\left(f_{x}\frac{X}{Z}+c_{x},\; f_{y}\frac{Y}{Z}+c_{y}\right)=(u,v),
\label{eq:projection}
\end{equation}
where $(f_{x},f_{y},c_{x},c_{y})$ are the camera intrinsic parameters.
Given a depth map $D(u,v)$, a 2D point $(u,v)$ backprojects to 3D via:
\begin{equation}
\pi^{-1}(u,v,D(u,v))=D(u,v)\!\left(\frac{u-c_{x}}{f_{x}},\frac{v-c_{y}}{f_{y}},1\right)^{\!\top}=(X,Y,Z).
\label{eq:backprojection}
\end{equation}
 
\subsubsection{Differentiable Warping}
 
The corresponding pixel location in source frame $I_{t}$ is computed by composing
backprojection, pose transformation, and projection:
\begin{equation}
(u',v')=\pi\!\left(M_{t}\,\pi^{-1}(u,v,D(u,v))\right)=g(u,v\mid D(u,v),M_{t}).
\label{eq:warping}
\end{equation}
Since $(u',v')$ are continuous-valued, we apply differentiable bilinear sampling
as in Spatial Transformer Networks~\cite{jaderberg2015spatial}:
\begin{equation}
I^{s}(u,v)=\sum_{i}\sum_{j}w^{ij}\,I_{t}(u',v'),
\label{eq:bilinear}
\end{equation}
where $w^{ij}$ is proportional to the spatial proximity between $(u,v)$ and $(u',v')$,
and $\sum_{i,j}w^{ij}=1$.

\subsection{JEPADepth: JEPA-Regularized Self-Supervised Depth Training}

We now describe \textbf{JEPADepth}, our approach for integrating an I-JEPA objective into a standard self-supervised monocular depth estimation framework. The core idea is to complement pixel-space photometric supervision with a 
\emph{masked prediction loss in representation space}, computed over patch-level 
tokens produced by a pretrained DINOv3 encoder. We hypothesize that this additional objective anchors the encoder to its pretrained feature manifold during photometric 
adaptation, preventing fine-tuning from eroding the cross-domain structure that 
DINOv3 pretraining encodes.

\paragraph{Patch-token representation backbone.}
Given an input image $I \in \mathbb{R}^{H \times W \times 3}$, a Vision Transformer encoder produces a sequence of patch tokens. We use a pretrained DINOv3 ViT and extract patch-wise features
\begin{equation}
\mathbf{Z} = E(I) \in \mathbb{R}^{N \times d},
\end{equation}
where $N = \frac{H}{P}\times\frac{W}{P}$ is the number of patches for patch size $P$, and $d$ is the embedding dimension. Unlike pixel reconstruction, these tokens capture higher-level structure such as object boundaries and layout, which we leverage as the feature space for predictive learning.

\paragraph{Depth decoder.}
To predict depth, we attach a lightweight convolutional decoder $D_{\omega}$ to the encoder features.
Given the patch-token embeddings $\mathbf{Z}$, the decoder upsamples and fuses multi-scale features to produce a disparity map $\hat{d}$:
\begin{equation}
\hat{d} = D_{\omega}(\mathbf{Z}).
\end{equation}
We follow the standard design used in self-supervised monocular depth estimation (e.g., \cite{godard2019digging}), using lateral projections and progressive upsampling to output disparities at multiple scales, which are then used by the photometric view-synthesis loss.

\begin{table}[H]
\centering
\caption{\textbf{FPN Decoder architecture.} $C_{\text{in}}$ is the ViT embedding
dimension ($384$ for ViT-S/16), $C=256$ is the decoder channel width, and
$H\!\times\!W$ is the input resolution. All feature maps entering the decoder
are at stride 16 (ViT patch grid).}
\label{tab:fpn-decoder}
\setlength{\tabcolsep}{10pt}  
\renewcommand{\arraystretch}{1.4} 
\resizebox{\columnwidth}{!}{%
\begin{tabular}{llccc}
\toprule
\textbf{Stage} & \textbf{Operation} & \textbf{Spatial size} & \textbf{Channels} & \textbf{Output} \\
\midrule
Lateral $\times4$ & $1\!\times\!1$ Conv (no bias) &
  $\frac{H}{16}\!\times\!\frac{W}{16}$ & $C_{\text{in}}\to C$ & \texttt{lat[0..3]} \\
\midrule
Stage 3 & Upsample $\times2$ (nearest) &
  $\frac{H}{16}\!\times\!\frac{W}{16}\to\frac{H}{8}\!\times\!\frac{W}{8}$ & $C$ & — \\
        & Add \texttt{lat[-2]} &
  $\frac{H}{8}\!\times\!\frac{W}{8}$ & $C$ & — \\
        & Conv-BN-ReLU ($3\!\times\!3$, pad 1) &
  $\frac{H}{8}\!\times\!\frac{W}{8}$ & $C\to C$ & — \\
        & Conv ($3\!\times\!3$, pad 1) + Sigmoid &
  $\frac{H}{8}\!\times\!\frac{W}{8}$ & $C\to 1$ & $\hat{d}_{3}$ \\
\midrule
Stage 2 & Upsample $\times2$ (nearest) &
  $\frac{H}{8}\!\times\!\frac{W}{8}\to\frac{H}{4}\!\times\!\frac{W}{4}$ & $C$ & — \\
        & Add \texttt{lat[-3]} &
  $\frac{H}{4}\!\times\!\frac{W}{4}$ & $C$ & — \\
        & Conv-BN-ReLU ($3\!\times\!3$, pad 1) &
  $\frac{H}{4}\!\times\!\frac{W}{4}$ & $C\to C$ & — \\
        & Conv ($3\!\times\!3$, pad 1) + Sigmoid &
  $\frac{H}{4}\!\times\!\frac{W}{4}$ & $C\to 1$ & $\hat{d}_{2}$ \\
\midrule
Stage 1 & Upsample $\times2$ (nearest) &
  $\frac{H}{4}\!\times\!\frac{W}{4}\to\frac{H}{2}\!\times\!\frac{W}{2}$ & $C$ & — \\
        & Add \texttt{lat[-4]} &
  $\frac{H}{2}\!\times\!\frac{W}{2}$ & $C$ & — \\
        & Conv-BN-ReLU ($3\!\times\!3$, pad 1) &
  $\frac{H}{2}\!\times\!\frac{W}{2}$ & $C\to C$ & — \\
        & Conv ($3\!\times\!3$, pad 1) + Sigmoid &
  $\frac{H}{2}\!\times\!\frac{W}{2}$ & $C\to 1$ & $\hat{d}_{1}$ \\
\midrule
Stage 0 & Upsample $\times2$ (nearest) &
  $\frac{H}{2}\!\times\!\frac{W}{2}\to H\!\times\!W$ & $C$ & — \\
        & Conv-BN-ReLU ($3\!\times\!3$, pad 1) &
  $H\!\times\!W$ & $C\to C$ & — \\
        & Conv ($3\!\times\!3$, pad 1) + Sigmoid &
  $H\!\times\!W$ & $C\to 1$ & $\hat{d}_{0}$ \\
\midrule
Outputs & $\hat{d}_{0}$: full resolution  & $H\!\times\!W$ & 1 & scale 0 \\
        & $\hat{d}_{1}$: half resolution  & $\frac{H}{2}\!\times\!\frac{W}{2}$ & 1 & scale 1 \\
        & $\hat{d}_{2}$: quarter resolution & $\frac{H}{4}\!\times\!\frac{W}{4}$ & 1 & scale 2 \\
        & $\hat{d}_{3}$: eighth resolution  & $\frac{H}{8}\!\times\!\frac{W}{8}$ & 1 & scale 3 \\
\bottomrule
\end{tabular}}
\end{table}

The depth decoder $D_{\omega}$ is a lightweight Feature Pyramid Network (FPN)-style
convolutional decoder that upsamples the patch-level ViT features to full image
resolution. All ViT features enter the decoder at stride 16 (the ViT patch grid).
The decoder produces multi-scale disparity outputs $\{\hat{d}_{0},\hat{d}_{1},\hat{d}_{2},\hat{d}_{3}\}$,
which are used by the photometric view-synthesis loss at each scale.
Table~\ref{tab:fpn-decoder} provides the complete architectural specification.
 
The design follows standard FPN principles: lateral $1\times1$ convolutions
project ViT tokens from embedding dimension $C_{\text{in}}=384$ (ViT-S/16) to a
uniform channel width $C=256$, followed by iterative top-down upsampling with
residual addition. Each stage outputs a single-channel disparity prediction via a
$3\times3$ convolution with sigmoid activation. The total decoder has approximately
$2$M parameters, making it lightweight relative to the $22$M encoder.

\begin{figure}[tb]
  \centering
  \includegraphics[height=3.3cm]{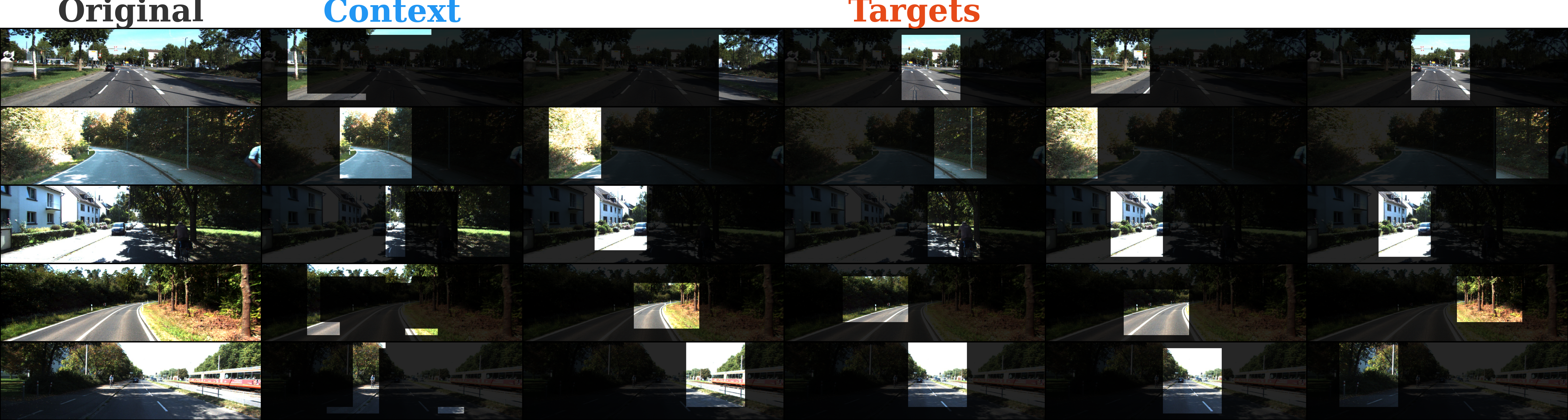}
  \caption{\textbf{Context-target masking strategy.}
For each input image, we randomly sample four target regions with scales drawn from the interval $(0.15, 0.2)$ and aspect ratios from $(0.75, 1.5)$. Subsequently, we sample a context region with scale in $(0.85, 1.0)$ and discard any overlapping target regions. This design encourages the target regions to capture semantically meaningful content, while the context region remains informative yet spatially sparse, ensuring computational efficiency.
  }
  \label{fig:context-masking-strategy}
\end{figure}

\paragraph{Context/target masking and JEPA prediction.}
Training follows the JEPA principle: predicting representations of \emph{masked target regions} from \emph{visible context regions}, rather than reconstructing pixels. For each image, we sample (i) a context mask $M_{ctx}$ selecting a subset of visible patches and (ii) one or more target masks $\{M_{tgt}^{(k)}\}_{k=1}^{K}$ selecting held-out patch regions. Masks are defined on the patch grid and can be sampled with varying scale and aspect ratio, leading to diverse spatial prediction tasks. Figure~\ref{fig:context-masking-strategy} shows examples of context and target blocks sampled in practice.

Let $\mathbf{Z}_{ctx} = \mathbf{Z}[M_{ctx}] \in \mathbb{R}^{N_{ctx} \times d}$ denote the context tokens. A predictor network $P_\psi$ takes these context embeddings and produces predictions for each target block:
\begin{equation}
\widehat{\mathbf{Z}}_{tgt}^{(k)} = P_\psi(\mathbf{Z}_{ctx}, M_{ctx}, M_{tgt}^{(k)}) \in \mathbb{R}^{N_{tgt}^{(k)} \times d}.
\end{equation}
Following common I-JEPA implementations, $P_\psi$ is a transformer that (i) embeds the context tokens into the predictor space, (ii) concatenates learned mask tokens for the target positions with fixed sinusoidal positional embeddings, and (iii) outputs predicted target embeddings projected back to the encoder dimension. This design allows the predictor to reason over the visible region while explicitly representing the missing target locations.

\paragraph{Target encoder and stop-gradient training.}
To stabilize training and avoid representational collapse, we employ a \emph{target encoder} $E'$ updated as an EMA of the context encoder $E$. The target encoder processes the image and provides target-region embeddings:
\begin{equation}
\mathbf{Z}' = E'(I), \qquad \mathbf{Z}_{tgt}^{\prime (k)}=\mathbf{Z}'[M_{tgt}^{(k)}].
\end{equation}
Gradients are stopped through the target branch, and only the context encoder and predictor are updated by backpropagation through the JEPA objective:
\begin{equation}
L_{\mathrm{JEPA}}
=
\frac{1}{K}\sum_{k=1}^{K}
\operatorname{L2}\!\left(
\widehat{\mathbf{Z}}_{tgt}^{(k)},
\operatorname{sg}\!\left(\mathbf{Z}_{tgt}^{\prime (k)}\right)
\right),
\label{eq:jepa_loss}
\end{equation}
where $\operatorname{sg}(\cdot)$ denotes the stop-gradient operator. The target encoder parameters are updated by EMA:
\begin{equation}
\phi \leftarrow m \phi + (1-m)\theta,
\end{equation}
where $\theta$ and $\phi$ denote the parameters of $E$ and $E'$, respectively, and $m$ is a momentum coefficient.

\subsection{Loss Function}
\textbf{Objective functions.} In line with the methodologies described
in \cite{godard2017unsupervised,godard2019digging}, we adopt the conventional photometric loss
$pe$, which is a combination of $L1$ and $SSIM$ losses:
\begin{equation}
pe(I_{a},I_{b})=\frac{\alpha}{2}(1-SSIM(I_{a},I_{b}))+(1-\alpha)\left\Vert I_{a}-I_{b}\right\Vert _{1}.\label{eq:important-7}
\end{equation}

To ensure proper depth regularization in areas lacking texture, we
employ an edge-aware smooth loss, applied in the following manner:
\begin{equation}
L_{s}=\left|\partial_{x}d_{t}^{*}\right|e^{-\left|\partial_{x}I_{t}\right|}+\left|\partial_{y}d_{t}^{*}\right|e^{-\left|\partial_{y}I_{t}\right|}, \label{eq:important-8}
\end{equation} where $\hat d_{t}^{*} = \hat d_t / \overline{\hat d_t}$ is the mean-normalized inverse depth,
following~\cite{godard2019digging}, to discourage shrinking of the estimated
depth.

\textbf{Masking Strategy. }In real-world settings, scenarios featuring
stationary cameras and moving objects can disrupt the usual assumptions
of a moving camera and static environment, negatively impacting the
performance of self-supervised depth estimators. We adopt the auto-masking strategy outlined in \cite{godard2019digging},
which filters out static pixels and areas of low texture that appear
unchanged between two consecutive frames in a sequence. This binary
mask $\mu$ is calculated as per \eqref{eq:important-9}, employing
the Iverson bracket notation:
\begin{equation}
\mu=[\underset{t'}{\mathrm{min}}\,pe(I_{t},I_{t'\to t})<\underset{t'}{\mathrm{min}}\,pe(I_{t},I_{t'})].\label{eq:important-9}
\end{equation}

The per-pixel photometric loss is taken as the minimum over source frames 
to handle occlusions~\cite{godard2019digging}:
\begin{equation}
L_p = \min_{t'} pe(I_t, I_{t' \to t}).
\end{equation}

\textbf{Final Training Loss.} We formulate the final loss by combining
the masked photometric objective, the edge-aware smoothness regularizer,
and the JEPA representation prediction objective. The JEPA term
$L_{\mathrm{JEPA}}$ is defined in Eq.~\eqref{eq:jepa_loss}:
\begin{equation}
L=\mu L_{p}+\lambda_{\text{smooth}} L_{s}+\lambda_{\mathrm{JEPA}}L_{\mathrm{JEPA}},
\label{eq:important-10}
\end{equation}
where $\lambda_{\text{smooth}}$ weights the smoothness term and $\lambda_{\mathrm{JEPA}}$ controls the contribution of the JEPA term.

\section{Experiments}
\label{sec:experiments}
We evaluate the effectiveness of JEPADepth on the KITTI, 
Cityscapes, and Make3D benchmarks. Model performance is quantified using the standard depth estimation metrics introduced in \cite{eigen2015predicting}, enabling direct comparison with prior work.

\subsection{Evaluation Metrics}
\label{sec:metrics}

We use the seven standard depth estimation metrics of~\cite{eigen2015predicting}
to quantify performance. The four error metrics are:

\paragraph{Absolute Relative Error (AbsRel):}
\begin{equation}
\frac{1}{n}\sum_{p}\frac{|y_p - \hat{y}_p|}{y_p}.
\end{equation}

\paragraph{Root Mean Squared Error (RMSE):}
\begin{equation}
\sqrt{\frac{1}{n}\sum_{p}(y_p - \hat{y}_p)^2}.
\end{equation}

\paragraph{Squared Relative Difference (SqRel):}
\begin{equation}
\frac{1}{n}\sum_{p}\frac{\|y_p - \hat{y}_p\|^2}{y_p}.
\end{equation}

\paragraph{RMSE log:}
\begin{equation}
\sqrt{\frac{1}{n}\sum_{p}\|\log y_p - \log \hat{y}_p\|^2}.
\end{equation}

The three threshold accuracy metrics $\delta_i$ measure the percentage of
pixels $y_p$ satisfying:
\begin{equation}
\max_{p}\!\left(\frac{y_p}{\hat{y}_p}, \frac{\hat{y}_p}{y_p}\right) = \delta < \text{thr},
\end{equation}
for $\text{thr} = 1.25,\ 1.25^2,\ 1.25^3$, where $y_p$ is a pixel in the
ground-truth depth map $y$, $\hat{y}_p$ is the corresponding pixel in the
predicted depth map $\hat{y}$, and $n$ is the total number of pixels per
depth map. Higher $\delta_i$ is better; lower is better for all error metrics.
 
\subsection{Datasets and Experimental Protocol}
 
\subsubsection{KITTI} \cite{geiger2013vision}
We evaluate JEPADepth on the widely adopted KITTI benchmark for self-supervised monocular depth estimation. Following standard practice, we use the Eigen split \cite{eigen2015predicting}, which consists of 39810 monocular triplets for training and
4424 for validation. Our model uses a pretrained DINOv3 encoder and is fine-tuned 
on KITTI under a strictly monocular self-supervised setting. Specifically, training relies solely on photometric reprojection loss with auto-masking \cite{godard2019digging}, without using stereo pairs, ground-truth depth supervision, or auxiliary datasets.  To compare with the existing solutions, we evaluate the single-view depth performance on the test
split of \cite{eigen2014depth} either using raw LiDAR (697 images) or improved ground truth labels \cite{uhrig2017sparsity} (652 images).
 
\begin{figure}[tb]
  \centering
  \includegraphics[height=8cm]{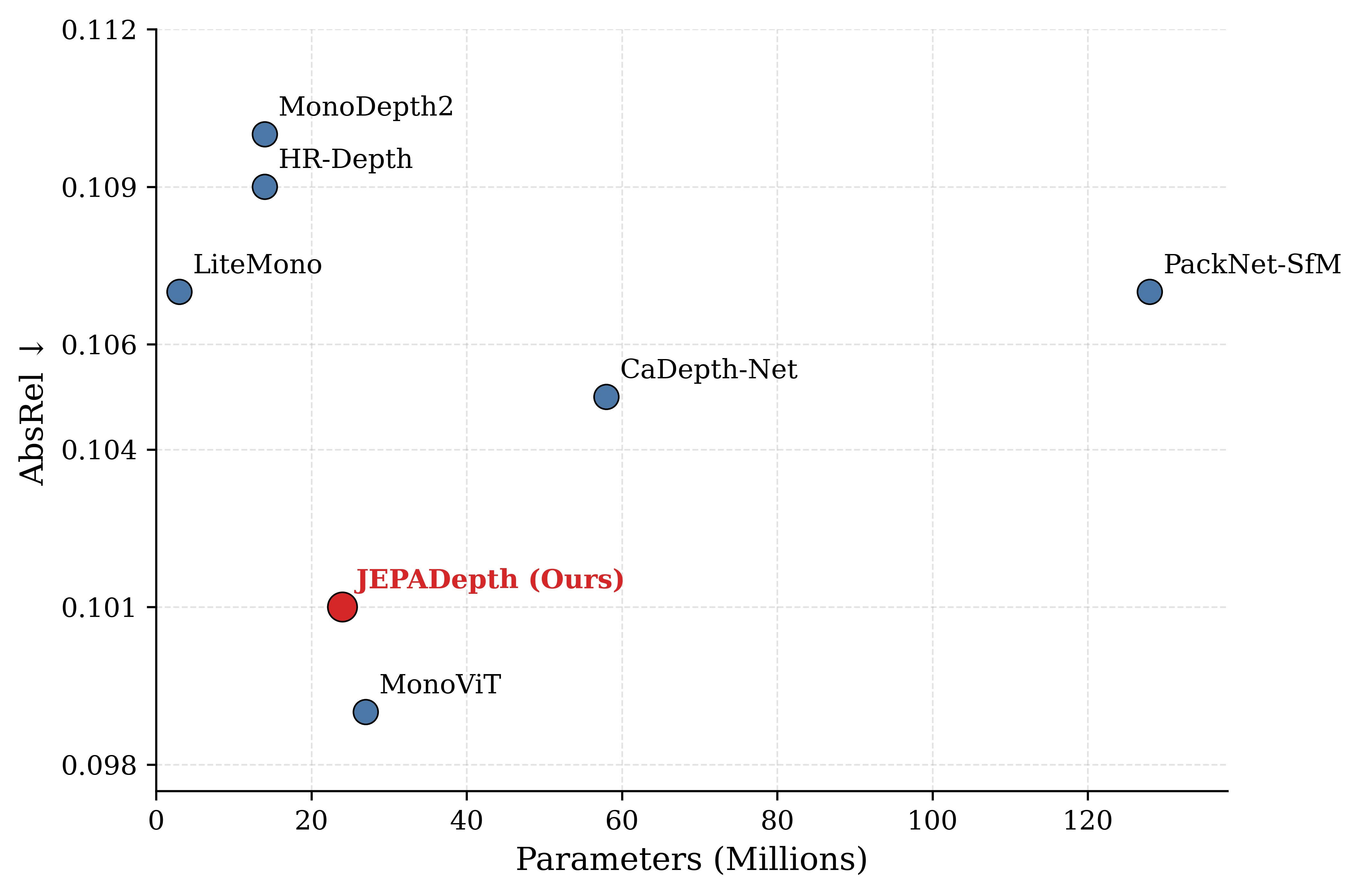}
  \caption{\textbf{Comparison of model efficiency and depth accuracy on KITTI using parameter count and AbsRel error}. Each point corresponds to a different self-supervised monocular depth method. The plot highlights the trade-off between model size and performance, showing that \textbf{JEPADepth} achieves competitive accuracy with a relatively compact model, indicating a favorable accuracy-efficiency balance compared to prior approaches.
  }
  \label{fig:params-performance-comparison}
\end{figure}
 
\subsubsection{Cityscapes} \cite{cordts2016cityscapes}
To assess cross-dataset generalization, we perform zero-shot evaluation on the Cityscapes dataset. Cityscapes is characterized by complex urban environments and a high density of dynamic objects, making it a challenging benchmark for generalization. We evaluate a model pretrained on KITTI directly on Cityscapes without any fine-tuning.  For fair comparison with prior work, we adopt the preprocessing protocol described in \cite{zhou2017unsupervised}, which converts image sequences into triplets and applies consistent cropping and resizing procedures. We report results on the 1525-image test split using the provided SGM \cite{hirschmuller2007stereo} disparity maps. As in the KITTI evaluation, predicted depths are clipped at 80m, and metrics are computed only on ground‑truth depth values below this threshold.
 
\subsubsection{Make3D} \cite{saxena2007learning}
To further analyze robustness and generalization to unseen environments, we conduct a zero-shot evaluation on Make3D. The model trained exclusively on KITTI is directly evaluated on Make3D without fine-tuning following the same image preprocessing steps
and computing the evaluation metrics as detailed in \cite{godard2019digging}.
 
\subsection{Implementation Details}
Our method is implemented in PyTorch \cite{paszke2019pytorch}.
Unless stated otherwise, all models are trained on the KITTI dataset using the Eigen split at a resolution of $192\times640$.
We apply standard data augmentations consisting of random horizontal flips and color jitter, following common practice in self-supervised monocular depth estimation \cite{godard2019digging}.
For optimization, we use Adam \cite{kingma2014adam} with $\beta_1=0.9$ and $\beta_2=0.999$, a base learning rate of $1\times10^{-4}$, and weight decay of $1\times10^{-2}$.
Training is run for $20$ epochs with a step learning-rate schedule that decays the learning rate by a factor of $0.1$ after epoch $15$ (i.e., to $1\times10^{-5}$ thereafter).
We employ SSIM-based photometric reconstruction with $\alpha=0.85$ (as in prior work \cite{godard2019digging}) and set the disparity smoothness regularization weight to $\lambda_{\text{smooth}}=10^{-3}$.
The training batch size is $12$ and all experiments are conducted on a single NVIDIA RTX 4090 GPU.
 
\begin{table*}[t]
  \caption{KITTI (Eigen split) depth estimation results. Lower is better ($\downarrow$), higher is better ($\uparrow$). $\delta_i$ denotes $\delta < 1.25^i$.
  Train: M=Monocular, MS=Monocular+Stereo, S=Stereo. All monocular methods use median scaling at test time. \textbf{Bold} indicates best result, \underline{underline} indicates second best.}
  \label{tab:kitti_eigen}
  \centering
  \small
  \setlength{\tabcolsep}{3pt}
  \renewcommand{\arraystretch}{1.1}
  \resizebox{\textwidth}{!}{%
  \begin{tabular}{@{}lcccccccccc@{}}
    \toprule
    Method & Train & Params & H$\times$W &
    AbsRel$\downarrow$ & SqRel$\downarrow$ & RMSE$\downarrow$ & RMSE$_{\text{log}}\downarrow$ &
    $\delta_1\uparrow$ & $\delta_2\uparrow$ & $\delta_3\uparrow$ \\
    \midrule
    \midrule
    Monodepth2~\cite{godard2019digging}      & M  & 14M  & $192\times640$ & 0.115 & 0.903 & 4.863 & 0.193 & 0.877 & 0.959 & 0.981 \\
    Monodepth2~\cite{godard2019digging}      & MS  & 14M  & $192\times640$ & 0.106 & 0.818 & 4.750 & 0.196 & 0.874 & 0.957 & 0.979 \\
    Depth Hints~\cite{watson2019self}      & S  & 34M  & $192\times640$ & 0.102 & \underline{0.762} & 4.602 & 0.189 & 0.880 & 0.960 & 0.981 \\
    CADepth-Net~\cite{yan2021channel}     & M  & 58M  & $192\times640$ & 0.105 & 0.769 & 4.535 & 0.181 & 0.892 & \underline{0.964} & \underline{0.983} \\
    PackNet-SfM~\cite{guizilini2019packnet}     & M  & 128M & $192\times640$ & 0.111 & 0.785 & 4.601 & 0.189 & 0.878 & 0.960 & 0.982 \\
    Johnston \textit{et al.}~\cite{johnston2020self}      & M  & 51M  & $192\times640$ & 0.106 & 0.861 & 4.699 & 0.185 & 0.889 & 0.962 & 0.982 \\
    HR-Depth~\cite{lyu2021hr}        & M  & 14M  & $192\times640$ & 0.109 & 0.792 & 4.632 & 0.185 & 0.884 & 0.962 & \underline{0.983} \\
    Lite-Mono~\cite{zhang2023lite}       & M  & 3M   & $192\times640$ & 0.107 & 0.765 & 4.561 & 0.183 & 0.886 & 0.963 & \underline{0.983} \\
    DIFFNet~\cite{zhou2021self}         & M  & 11M  & $192\times640$ & 0.102 & 0.764 & \underline{4.483} & \underline{0.180} & 0.896 & \underline{0.965} & \underline{0.983} \\
    MonoViT~\cite{zhao2022monovit}         & M  & 27M  & $192\times640$ & \textbf{0.099} & \textbf{0.708} & \textbf{4.372} & \textbf{0.175} & \textbf{0.900} & \textbf{0.967} & \textbf{0.984} \\
    \midrule
    DINOv3 w/o JEPA & M  & 24M  & $192\times640$ & 0.105 & 0.861 & 4.710 & 0.186 & 0.891 & 0.961 & 0.981 \\
    JEPADepth (Ours)             & M  & 24M  & $192\times640$ & \underline{0.101} & 0.782 & 4.563 & \underline{0.180} & \underline{0.898} & \underline{0.965} & \underline{0.983} \\
    \bottomrule
  \end{tabular}
  }
\end{table*}
 
\begin{table*}[t]
  \caption{\textbf{Performance comparison on KITTI improved ground truth}~\cite{uhrig2017sparsity}. 
  Lower is better ($\downarrow$), higher is better ($\uparrow$). $\delta_i$ denotes $\delta < 1.25^i$. Train: M=Monocular. All monocular methods use median scaling at test time. \textbf{Bold} indicates best result, \underline{underline} indicates second best.}
  \label{tab:kitti_improved}
  \centering
  \small
  \setlength{\tabcolsep}{3pt}
  \renewcommand{\arraystretch}{1.1}
  \resizebox{\textwidth}{!}{%
  \begin{tabular}{@{}lcccccccccc@{}}
    \toprule
    Method & Train & Params & H$\times$W &
    AbsRel$\downarrow$ & SqRel$\downarrow$ & RMSE$\downarrow$ & RMSE$_{\text{log}}\downarrow$ &
    $\delta_1\uparrow$ & $\delta_2\uparrow$ & $\delta_3\uparrow$ \\
    \midrule
    \midrule
    Monodepth2~\cite{godard2019digging}    & M  & 14M  & $192\times640$ & 0.090 & 0.545 & 3.942 & 0.137 & 0.914 & 0.983 & 0.995 \\
    HR-Depth~\cite{lyu2021hr}      & M  & 14M  & $192\times640$ & 0.085 & 0.471 & 3.769 & 0.130 & 0.919 & 0.985 & \underline{0.996} \\
    PackNet-SfM~\cite{guizilini2019packnet}   & M  & 128M & $192\times640$ & 0.078 & 0.420 & \underline{3.485} & 0.121 & 0.931 & 0.986 & \underline{0.996} \\
    CADepth-Net~\cite{yan2021channel}   & M  & 58M  & $192\times640$ & 0.080 & 0.442 & 3.639 & 0.124 & 0.927 & 0.986 & \underline{0.996} \\
    DIFFNet~\cite{zhou2021self}       & M  & 11M   & $192\times640$ & \underline{0.076} & \underline{0.412} & 3.494 & \underline{0.119} & \underline{0.935} & \underline{0.988} & \underline{0.996} \\
    MonoViT~\cite{zhao2022monovit}       & M  & 27M  & $192\times640$ & \textbf{0.075} & \textbf{0.389} & \textbf{3.419} & \textbf{0.115} & \textbf{0.938} & \textbf{0.989} & \textbf{0.997} \\
    \midrule
    DINOv3 w/o JEPA & M  & 24M  & $192\times640$ & 0.077 & 0.476 & 3.724 & 0.123 & 0.932 & 0.985 & 0.995 \\
    JEPADepth (Ours)           & M  & 24M  & $192\times640$ & \textbf{0.075} & 0.446 & 3.621 & 0.120 & \underline{0.935} & 0.987 & \underline{0.996} \\
 
    \bottomrule
  \end{tabular}
  }
\end{table*}
 
\begin{table*}[t]
  \caption{\textbf{Cityscapes depth estimation results.}
  Lower is better ($\downarrow$), higher is better ($\uparrow$). $\delta_i$ denotes $\delta < 1.25^i$. Train: K=KITTI, C=Cityscapes. All monocular methods use median scaling at test time. \textbf{Bold} indicates best result, \underline{underline} indicates second best.}
  \label{tab:cityscapes}
  \centering
  \small
  \setlength{\tabcolsep}{3pt}
  \renewcommand{\arraystretch}{1.1}
  \resizebox{\textwidth}{!}{%
  \begin{tabular}{@{}lcccccccccc@{}}
    \toprule
    Method & Train &
    AbsRel$\downarrow$ & SqRel$\downarrow$ & RMSE$\downarrow$ & RMSE$_{\text{log}}\downarrow$ &
    $\delta_1\uparrow$ & $\delta_2\uparrow$ & $\delta_3\uparrow$ \\
    \midrule
    Pilzer \textit{et al.}~\cite{pilzer2018unsupervised}    & C & 0.240 & 4.264 & 8.049 & 0.334 & 0.710 & 0.871 & 0.937 \\
    Monodepth2 ~\cite{godard2019digging}    & K & 0.163 & 1.883 & 8.967 & 0.241 & 0.757 & 0.922 & 0.974 \\
    CADepth-Net ~\cite{yan2021channel}    & K & 0.150 & 1.691 & 8.527 & 0.227 & 0.786 & 0.931 & 0.978 \\
    Lite-Mono ~\cite{zhang2023lite}    & K & 0.153 & 1.636 & 8.390 & 0.225 & 0.782 & 0.933 & 0.979 \\
    DIFFNet ~\cite{zhou2021self}    & K & \underline{0.146} & 1.610 & 8.188 & 0.218 & 0.799 & 0.938 & 0.980 \\
    MonoViT ~\cite{zhao2022monovit}    & K & \textbf{0.143} & \textbf{1.485} & \underline{7.981} & \underline{0.211} & \underline{0.803} & \underline{0.942} & \underline{0.983} \\
    JEPADepth (Ours)           & K  & \textbf{0.143} & \underline{1.495} & \textbf{7.766} & \textbf{0.208} & \textbf{0.810} & \textbf{0.948}& \textbf{0.985}\\
    \bottomrule
  \end{tabular}
  }
\end{table*}
 
\begin{table}[t]
  \caption{\textbf{Make3D depth estimation results.}
  Lower is better ($\downarrow$). \textbf{Bold} indicates best result, \underline{underline} indicates second best.}
  \label{tab:make3d}
  \centering
  \small
  \setlength{\tabcolsep}{5pt}
  \renewcommand{\arraystretch}{1.1}
  \begin{tabular}{@{}lcccc@{}}
    \toprule
    Method & AbsRel$\downarrow$ & SqRel$\downarrow$ & RMSE$\downarrow$ & RMSE$_{\text{log}}\downarrow$ \\
    \midrule
    Zhou~\cite{zhou2017unsupervised}        & 0.383 & 5.321 & 10.470 & 0.478 \\
    DDVO~\cite{wang2018learning}        & 0.387 & 4.720 & 8.090  & 0.204 \\
    Monodepth2~\cite{godard2019digging}  & 0.322 & 3.589 & 7.417  & 0.163 \\
    CADepth-Net~\cite{yan2021channel} & 0.312 & 3.086 & 7.066  & 0.159 \\
    HR-Depth~\cite{lyu2021hr} & 0.305 & 2.944 & 6.857  & 0.157 \\
    Lite-Mono~\cite{zhang2023lite} & 0.305 & 3.060 & 6.981  & 0.158 \\
    DIFFNet ~\cite{zhou2021self} & 0.298 & 2.901 & 6.753  & 0.153 \\
    MonoViT~\cite{zhao2022monovit} & 0.286 & \textbf{2.758} & \textbf{6.623}  & \underline{0.147} \\
    \midrule
    DINOv3 w/o JEPA                      & \underline{0.284} & 3.135 & 6.903  & \underline{0.147} \\
    JEPADepth (Ours)         & \textbf{0.275} & \underline{2.825} & \underline{6.625} & \textbf{0.143} \\
    \bottomrule
  \end{tabular}
\end{table}
 
\paragraph{JEPA training configuration.}
When enabling JEPA training, the depth model is built around a DINOv3 ViT-S/16 encoder ($\sim$22M parameters) initialized from pretrained weights, coupled with a lightweight convolutional depth decoder ($\sim$2M parameters). The predictor is a 12-layer transformer with embedding dimension $384$ ($\sim$21M parameters) used exclusively during training; at inference time, only the context encoder and depth decoder are retained, yielding a $\sim$24M parameter inference model. The target encoder is likewise discarded at inference time.
In addition to the standard photometric objective, we include a JEPA-style masked representation prediction loss with weight $\lambda_{\text{JEPA}}=1$ selected based on a sensitivity analysis reported in Table~\ref{tab:ablation_lambda}, which shows that this value consistently yields the best results across all metrics.
Masks are generated on the patch grid using patch size $P=16$, following the masking configuration of the original I-JEPA work~\cite{assran2023self}, with one context mask per image covering $85$-$100\%$ of patches, a large coverage shown in~\cite{assran2023self} to be critical for providing sufficient spatial signal for semantic prediction, and four target masks covering $15$-$20\%$ of patches each; target masks are sampled with aspect ratio in $[0.75,1.5]$ and are constrained to be non-overlapping.
The target encoder is updated using an EMA of the context encoder parameters with momentum linearly scheduled from $0.996$ to $1.0$ over training. We use a linear schedule for simplicity; preliminary experiments showed no significant difference with cosine scheduling on KITTI.
 
The additional compute introduced by the JEPA component, specifically the extra forward pass through the target encoder and the predictor, results in a modest training overhead. Training JEPADepth on KITTI for 20 epochs takes approximately 8.6 hours on a single NVIDIA RTX 4090 GPU, compared to 7.4 hours for the photometric-only DINOv3 baseline, representing a $\sim$16\% increase in wall-clock training time. Critically, this overhead is incurred only during training; the predictor and target encoder are discarded at inference time, and the two models are identical in deployment cost.
 
\subsection{KITTI Results}
 
Tables~\ref{tab:kitti_eigen} and~\ref{tab:kitti_improved} report results on the KITTI 
Eigen split using standard and improved ground truth, respectively. JEPADepth achieves 
competitive performance among monocular methods, outperforming strong baselines such as 
Monodepth2, HR-Depth, PackNet-SfM, CADepth-Net, and Lite-Mono on both benchmarks. On 
the standard Eigen split, JEPADepth attains an AbsRel of $0.101$ and $\delta_1 = 0.898$, 
ranking second overall and surpassing all CNN-based methods while remaining competitive 
with MonoViT~\cite{zhao2022monovit}, which uses a larger hybrid transformer architecture. 
Notably, JEPADepth also achieves second-best performance on RMSE$_{\text{log}}$ 
($0.180$), $\delta_2$ ($0.965$), and $\delta_3$ ($0.983$) on the standard split. On 
the improved ground truth benchmark, JEPADepth ties MonoViT on AbsRel ($0.075$) while 
using fewer parameters ($24$M vs.\ $27$M), though MonoViT leads on the remaining metrics, and achieves second-best $\delta_1$ 
alongside DIFFNet ($0.935$). Furthermore, JEPADepth achieves this with a relatively compact 
model, as illustrated in Figure~\ref{fig:params-performance-comparison}, indicating a 
favorable accuracy-efficiency trade-off compared to heavier methods such as 
PackNet-SfM ($128$M) and CADepth-Net ($58$M).

\begin{figure}[H]
\centering
\includegraphics[width=\linewidth]{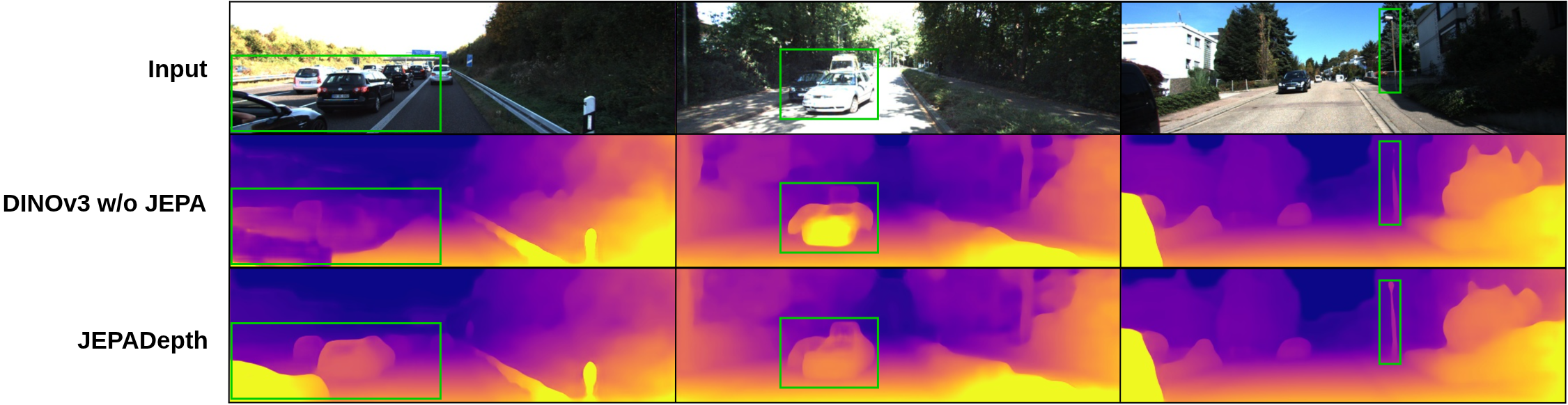}
\caption{\textbf{Qualitative KITTI results.} Depth predictions on the KITTI Eigen
split. JEPADepth (with the JEPA objective) recovers finer structure than the
photometric-only DINOv3 baseline.}
\label{fig:kitti-qual}
\end{figure}

Figure~\ref{fig:kitti-qual} shows qualitative depth predictions on the
KITTI Eigen split. The JEPA objective yields sharper object boundaries and more coherent scene structure than
the photometric-only DINOv3 baseline.
 
To place JEPADepth in the context of the most recent literature, Table~\ref{tab:recent} 
compares it against three contemporary self-supervised methods---DaCCN~\cite{han2023self}, 
HybridDepth~\cite{zhang2025hybrid}, and AdaDepth~\cite{gao2026adadepth}---using their 
published numbers. On the standard KITTI Eigen split, JEPADepth is competitive with 
DaCCN (AbsRel $0.101$ vs.\ $0.099$) and outperforms AdaDepth (AbsRel $0.101$ vs.\ 
$0.104$). HybridDepth attains the strongest KITTI scores, though it relies on language 
guidance and a coarse-to-fine module that remain active at inference time, whereas the 
JEPA predictor and target encoder in our method are discarded after training and add no 
inference cost. All three methods are evaluated against the same established baselines 
used throughout this work (e.g., Monodepth2, DIFFNet, Lite-Mono), confirming these remain 
common reference points. In cross-domain transfer, HybridDepth and AdaDepth report no 
zero-shot results and DaCCN reports only Make3D; on Make3D, JEPADepth outperforms DaCCN 
on both reported metrics (AbsRel $0.275$ vs.\ $0.290$; SqRel $2.825$ vs.\ $2.873$; 
cf.\ Table~\ref{tab:make3d}), consistent with our finding that the JEPA objective 
strengthens cross-domain generalization.
 
\begin{table}[H]
\centering
\setlength{\tabcolsep}{2.0pt}
\footnotesize
\caption{\textbf{KITTI Eigen and Make3D (zero-shot) comparison with recent
methods.} $\dagger$\,=\,published numbers. \textbf{Bold}\,=\,best,
\underline{underline}\,=\,second. $^*$Not reported.}
\label{tab:recent}
\begin{tabular}{lcccccc}
\toprule
& \multicolumn{4}{c}{KITTI Eigen} & \multicolumn{2}{c}{Make3D (0-shot)} \\
\cmidrule(lr){2-5}\cmidrule(lr){6-7}
Method & AbsRel$\downarrow$ & SqRel$\downarrow$ & RMSE$\downarrow$ & $\delta_1$$\uparrow$ & AbsRel$\downarrow$ & SqRel$\downarrow$ \\
\midrule
DaCCN~\cite{han2023self}$\dagger$          & \underline{0.099} & \textbf{0.661} & \textbf{4.316} & 0.897          & 0.290          & 2.873          \\
HybridDepth~\cite{zhang2025hybrid}$\dagger$ & \textbf{0.093}    & \underline{0.596} & \underline{4.113} & \textbf{0.910} & $^*$        & $^*$           \\
AdaDepth~\cite{gao2026adadepth}$\dagger$    & 0.104             & 0.749          & 4.491          & 0.891          & $^*$           & $^*$           \\
\textbf{JEPADepth (Ours)}                   & 0.101             & 0.782          & 4.563          & \underline{0.898} & \textbf{0.275} & \textbf{2.825} \\
\bottomrule
\end{tabular}
\end{table}

\subsection{Cityscapes Results}
 
Table~\ref{tab:cityscapes} reports zero-shot transfer results on Cityscapes, where 
JEPADepth is evaluated directly without any fine-tuning using only the model pretrained 
on KITTI. JEPADepth ties MonoViT for the best AbsRel ($0.143$) and 
achieves the best performance on RMSE ($7.766$), RMSE$_{\text{log}}$ ($0.208$), 
$\delta_1$ ($0.810$), $\delta_2$ ($0.948$), and $\delta_3$ ($0.985$), outperforming all competing methods on five out of seven metrics.
 
\begin{table}[t]
  \caption{\textbf{Ablation: effect of the JEPA objective and encoder fine-tuning.} Freezing the DINOv3 encoder and training only the decoder leads to 
  performance degradation, confirming that end-to-end fine-tuning is essential. 
  The JEPA loss further improves over fine-tuning alone. $\delta_i$ denotes $\delta < 1.25^i$.}
  \label{tab:ablation_jepa}
  \centering
  \small
  \setlength{\tabcolsep}{5pt}
  \renewcommand{\arraystretch}{1.1}
  \resizebox{\textwidth}{!}{%
  \begin{tabular}{@{}lcccccccc@{}}
    \toprule
    Method &
    AbsRel$\downarrow$ & SqRel$\downarrow$ & RMSE$\downarrow$ & RMSE$_{\text{log}}\downarrow$ &
    $\delta_1\uparrow$ & $\delta_2\uparrow$ & $\delta_3\uparrow$ \\
    \midrule
    DINOv3 frozen (decoder only) w/o JEPA  & 0.114 & 0.880 & 4.741 & 0.188 & 0.876 & 0.961 & 0.981 \\
    DINOv3 fine-tuned w/o JEPA & 0.105 & 0.861 & 4.710 & 0.186 & 0.891 & 0.961 & 0.981 \\
    DINOv3 fine-tuned w/ JEPA (Ours) & \textbf{0.101} & \textbf{0.782} & \textbf{4.563} & \textbf{0.180} & \textbf{0.898} & \textbf{0.965} & \textbf{0.983} \\
    \bottomrule
  \end{tabular}
  }
\end{table}
 
\begin{table}[t]
  \caption{\textbf{Ablation: effect of pretraining.} Without pretrained DINOv3 weights, the JEPA loss provides minimal benefit, highlighting the importance of strong initial representations. $\delta_i$ denotes $\delta < 1.25^i$.}
  \label{tab:ablation_pretrain}
  \centering
  \small
  \setlength{\tabcolsep}{5pt}
  \renewcommand{\arraystretch}{1.1}
  \resizebox{\textwidth}{!}{%
  \begin{tabular}{@{}lcccccccc@{}}
    \toprule
    Method &
    AbsRel$\downarrow$ & SqRel$\downarrow$ & RMSE$\downarrow$ & RMSE$_{\text{log}}\downarrow$ &
    $\delta_1\uparrow$ & $\delta_2\uparrow$ & $\delta_3\uparrow$ \\
    \midrule
    DINOv3 w/ JEPA, random init & 0.157 & 1.337 & 5.750 & 0.234 & 0.795 & 0.930 & 0.971 \\
    DINOv3 w/ JEPA, pretrained (Ours) & \textbf{0.101} & \textbf{0.782} & \textbf{4.563} & \textbf{0.180} & \textbf{0.898} & \textbf{0.965} & \textbf{0.983} \\
    \bottomrule
  \end{tabular}
  }
\end{table}
 
\begin{table}[t]
  \caption{\textbf{Ablation: prediction space of the JEPA loss.} Computing the JEPA objective in representation space outperforms pixel-space reconstruction, consistent with the I-JEPA design principle. $\delta_i$ denotes $\delta < 1.25^i$.}
  \label{tab:ablation_space}
  \centering
  \small
  \setlength{\tabcolsep}{5pt}
  \renewcommand{\arraystretch}{1.1}
  \resizebox{\textwidth}{!}{%
  \begin{tabular}{@{}lcccccccc@{}}
    \toprule
    Method &
    AbsRel$\downarrow$ & SqRel$\downarrow$ & RMSE$\downarrow$ & RMSE$_{\text{log}}\downarrow$ &
    $\delta_1\uparrow$ & $\delta_2\uparrow$ & $\delta_3\uparrow$ \\
    \midrule
    JEPA loss in pixel space   & 0.103 & 0.815 & 4.592 & 0.183 & 0.896 & 0.963 & 0.982 \\
    JEPA loss in repr.\ space (Ours) & \textbf{0.101} & \textbf{0.782} & \textbf{4.563} & \textbf{0.180} & \textbf{0.898} & \textbf{0.965} & \textbf{0.983} \\
    \bottomrule
  \end{tabular}
  }
\end{table}

\begin{table}[t]
  \caption{\textbf{Ablation: sensitivity to the JEPA loss weight $\lambda_{\text{JEPA}}$.} Results on KITTI Eigen split. $\lambda_{\text{JEPA}}=1.0$ achieves the best overall performance.}
  \label{tab:ablation_lambda}
  \centering
  \small
  \setlength{\tabcolsep}{5pt}
  \renewcommand{\arraystretch}{1.1}
  \begin{tabular}{@{}ccccccccc@{}}
    \toprule
    $\lambda_{\text{JEPA}}$ &
    AbsRel$\downarrow$ & SqRel$\downarrow$ & RMSE$\downarrow$ & RMSE$_{\text{log}}\downarrow$ &
    $\delta_1\uparrow$ & $\delta_2\uparrow$ & $\delta_3\uparrow$ \\
    \midrule
    0.6 & 0.108 & 0.839 & 4.705 & 0.187 & 0.885 & 0.961 & 0.981 \\
    0.8 & 0.110 & 0.852 & 4.796 & 0.190 & 0.879 & 0.958 & 0.981 \\
    \textbf{1.0} & \textbf{0.101} & \textbf{0.782} & \textbf{4.563} & \textbf{0.180} & \textbf{0.898} & \textbf{0.965} & \textbf{0.983} \\
    1.2 & 0.103 & 0.796 & 4.595 & 0.182 & 0.896 & 0.963 & 0.982 \\
    1.4 & 0.103 & 0.791 & 4.587 & 0.182 & 0.894 & 0.963 & 0.982 \\
    \bottomrule
  \end{tabular}
\end{table}
 
\subsection{Make3D Results}
Table~\ref{tab:make3d} reports zero-shot evaluation results on Make3D, where the model 
pretrained on KITTI is applied directly without any fine-tuning or domain adaptation. 
JEPADepth achieves the best AbsRel ($0.275$) and RMSE$_{\text{log}}$ 
($0.143$), outperforming all baselines including MonoViT~\cite{zhao2022monovit} 
on these metrics. MonoViT obtains better SqRel ($2.758$ vs.\ $2.825$) and marginally better RMSE ($6.623$ vs.\ $6.625$). The AbsRel improvement over MonoViT ($0.286 \rightarrow 0.275$) represents 
a relative gain of $3.8\%$, and over the strongest CNN baseline DIFFNet 
($0.298 \rightarrow 0.275$) a gain of $7.7\%$. 
 
Table~\ref{tab:make3d} also includes the DINOv3 baseline trained without JEPA 
under the same zero-shot protocol (AbsRel: $0.284$), confirming that the 
improvement over prior methods is not solely attributable to the DINOv3 encoder.

\begin{figure}[tb]
  \centering
  \includegraphics[height=10cm]{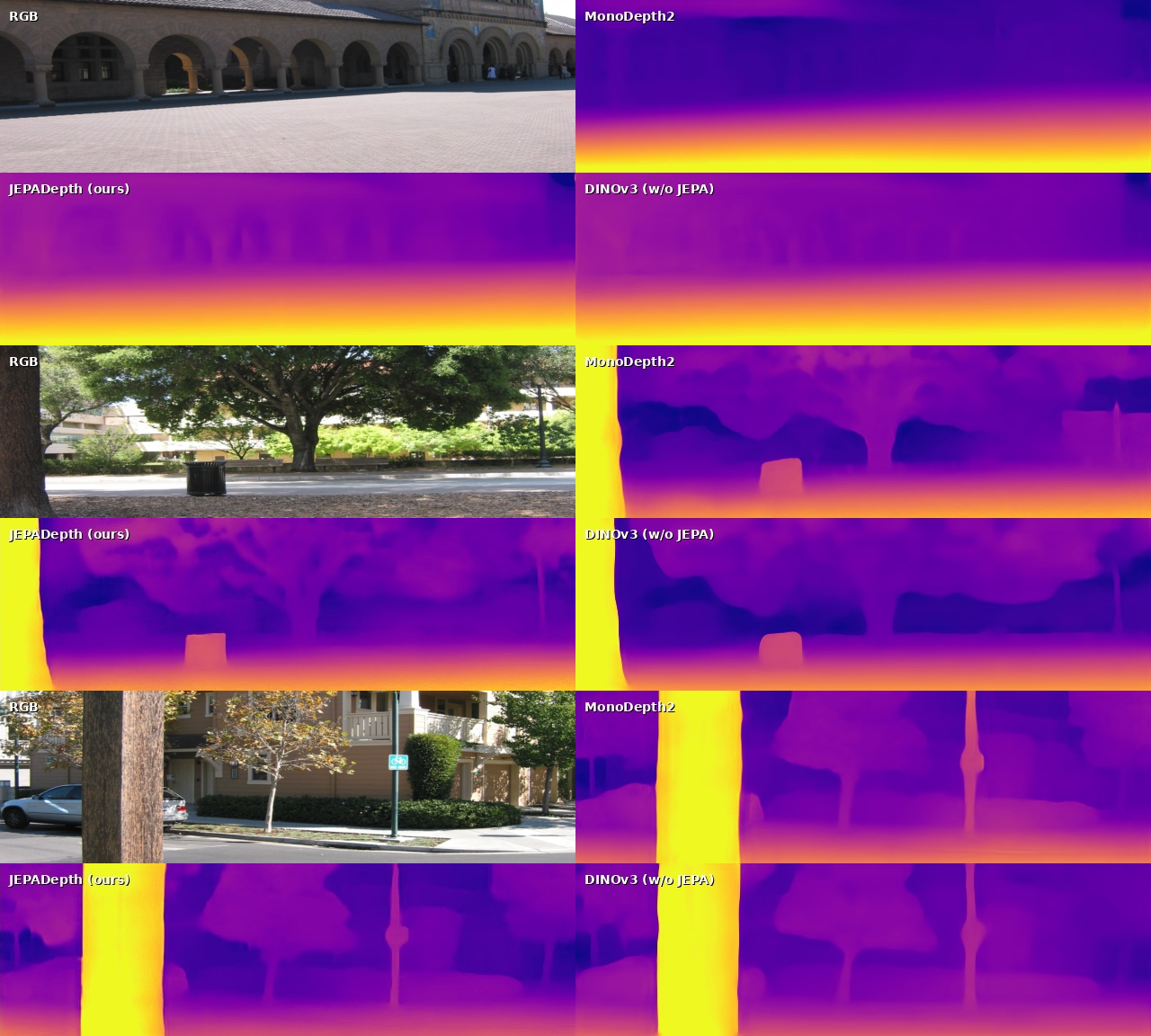}
  \caption{\textbf{Qualitative depth estimation results on Make3D (zero-shot).} Each 
  row shows, from top-left to bottom-right: input RGB image, Monodepth2 prediction, JEPADepth (Ours) prediction, and DINOv3 (w/o JEPA) prediction. All models are 
  evaluated zero-shot, trained exclusively on KITTI. JEPADepth produces 
  structurally coherent depth maps with well-defined object boundaries, correctly 
  separating foreground elements such as trees, poles, and buildings from the 
  background, whereas Monodepth2 and DINOv3 (w/o JEPA) tend to produce flatter, 
  less structured predictions that fail to recover fine-grained scene geometry under 
  domain shift.
  }
  \label{fig:qualitative_make3d}
\end{figure}

Qualitative results in 
Figure~\ref{fig:qualitative_make3d} further support these findings: JEPADepth produces 
depth maps with better scene-level understanding, correctly recovering the overall 
layout and relative depth ordering of outdoor structures even under significant domain 
shift from the KITTI training distribution. In contrast, the DINOv3 baseline without 
JEPA regularization tends to produce flatter, less structured predictions, suggesting 
that the masked predictive objective encourages the model to reason about spatial 
geometry rather than relying solely on local photometric cues.

\subsection{Ablation Study}

We conduct five ablation experiments to validate the key design choices of JEPADepth. All variants are evaluated on the KITTI Eigen split.
 
\paragraph{Effect of the JEPA objective and encoder fine-tuning.}
Table~\ref{tab:ablation_jepa} shows a three-way comparison motivating
JEPADepth. Freezing the DINOv3 encoder and training only the decoder yields
AbsRel $0.114$, a result comparable to basic CNN baselines such as
Monodepth2 ($0.115$), confirming that end-to-end adaptation is necessary
to go beyond this level. End-to-end photometric fine-tuning improves this to
AbsRel $0.105$. Adding the JEPA objective on top achieves the best result
(AbsRel: $0.101$), improving over the no-JEPA baseline consistently across all
seven metrics.
 
\paragraph{Effect of pretraining.}
Table~\ref{tab:ablation_pretrain} isolates the contribution of the pretrained 
DINOv3 weights. When the encoder is randomly initialized and trained with the 
JEPA objective, performance drops substantially (AbsRel: $0.157$), falling 
below even the fine-tuned no-JEPA baseline ($0.105$). This indicates that the 
benefits of JEPA regularization are tightly coupled with the quality of the 
pretrained representations: without rich initial features, the masked prediction 
loss cannot provide meaningful targets.
 
\paragraph{Sensitivity to the JEPA loss weight.}
Table~\ref{tab:ablation_lambda} reports the effect of varying $\lambda_{\text{JEPA}}$ on KITTI depth estimation performance. The model is robust to values in the range $[1.0, 1.4]$, with $\lambda_{\text{JEPA}}=1.0$ achieving the best scores. Performance degrades more noticeably for smaller weights ($\leq 0.8$), suggesting that insufficient JEPA regularization reduces its benefit, while slightly larger weights preserve most of the gain.

\paragraph{Pixel-space vs.\ representation-space prediction.}
Table~\ref{tab:ablation_space} ablates the prediction space of the JEPA loss. Replacing the latent-space regression with a pixel-space reconstruction loss (i.e., predicting raw pixel intensities for masked regions instead of feature-space embeddings) results in a modest but consistent performance drop (AbsRel: $0.103$ vs.\ $0.101$). While the gap is smaller than the other ablations, it confirms that operating in representation space, as in the original I-JEPA formulation, is preferable.

\paragraph{Decoder architecture: FPN vs.\ DPT.}
Table~\ref{tab:dpt} compares our lightweight FPN decoder against a DPT-style
decoder~\cite{ranftl2021vision}, each with and without the JEPA objective. The JEPA loss consistently
improves all metrics over the no-JEPA baseline across both decoders, confirming
that the benefit is decoder-agnostic and originates in the encoder. The DPT
decoder underperforms FPN under our self-supervised regime, as it was designed
for large-scale supervised settings.
 
\begin{table}[H]
\centering
\setlength{\tabcolsep}{3pt}
\footnotesize
\caption{\textbf{KITTI Eigen: FPN vs.\ DPT decoder, with and without JEPA.}
\textbf{Bold}\,=\,best.}
\label{tab:dpt}
\begin{tabular}{llcccc}
\toprule
Decoder & JEPA & AbsRel$\downarrow$ & SqRel$\downarrow$ & RMSE$\downarrow$ & $\delta_1$$\uparrow$ \\
\midrule
FPN & \ding{55} & 0.105 & 0.861 & 4.710 & 0.891 \\
FPN & \ding{51} & \textbf{0.101} & \textbf{0.782} & \textbf{4.563} & \textbf{0.898} \\
\midrule
DPT & \ding{55} & 0.109 & 0.908 & 4.792 & 0.884 \\
DPT & \ding{51} & 0.105 & 0.788 & 4.633 & 0.888 \\
\bottomrule
\end{tabular}
\end{table}

\section{Conclusion}
In this paper, we presented JEPADepth, a self-supervised monocular depth 
estimation framework that integrates an I-JEPA masked prediction objective 
in representation space alongside standard photometric supervision. Our 
ablations show that while a frozen DINOv3 encoder already matches basic CNN 
baselines, end-to-end fine-tuning with the JEPA objective consistently improves 
over photometric fine-tuning alone, across all metrics on KITTI and on 
zero-shot transfer to Make3D and Cityscapes, without any additional inference 
cost.

Experiments on KITTI demonstrate that JEPADepth achieves competitive performance among monocular self-supervised methods, outperforming all CNN-based baselines and approaching the best transformer-based competitor while using fewer parameters. Ablation studies confirm that each design choice contributes meaningfully: the JEPA objective, the pretrained DINOv3 backbone, and the representation-space prediction all play important roles in the final performance. Crucially, the benefits of JEPADepth extend beyond the training distribution. Zero-shot evaluations on Make3D and Cityscapes show consistent improvements: 
JEPADepth achieves the best AbsRel on Make3D and ties for the best AbsRel on 
Cityscapes while leading on the majority of metrics, without any fine-tuning.
The JEPA regularization introduces only a modest training overhead ($\sim$16\% over the photometric-only baseline) while yielding consistent improvements in depth accuracy and cross-dataset generalization, making it a practical addition to existing self-supervised depth pipelines.
A current limitation of JEPADepth is the relatively simple convolutional depth decoder ($\sim$2M parameters), which may bottleneck the rich representations produced by the DINOv3 encoder. Our ablation shows that naively substituting a more expressive DPT-style decoder does not improve results under the self-supervised regime (Table~\ref{tab:dpt}); designing a decoder that better exploits these representations without the large-scale supervision DPT was built for remains an open direction for future work.


%
%
\bibliographystyle{splncs04}
\bibliography{main}
\end{document}